\documentclass[10pt,twocolumn,letterpaper]{article}

\usepackage[pagenumbers]{cvpr} 

\usepackage{graphicx}
\usepackage{amsmath}
\usepackage{amssymb}
\usepackage{booktabs}
\usepackage{epigraph}

\usepackage{bm}
\usepackage{multirow}
\makeatletter
\newcommand{\rmnum}[1]{\romannumeral #1}
\newcommand{\Rmnum}[1]{\expandafter\@slowromancap\romannumeral #1@}
\makeatother
\usepackage[accsupp]{axessibility}

\usepackage[pagebackref,breaklinks,colorlinks,citecolor=cyan]{hyperref}

\usepackage[capitalize]{cleveref}
\crefname{section}{Sec.}{Secs.}
\Crefname{section}{Section}{Sections}
\Crefname{table}{Table}{Tables}
\crefname{table}{Tab.}{Tabs.}

\def\cvprPaperID{1} 
\def\confName{CVPR}
\def\confYear{2023}

\begin{document}

\title{Context De-confounded Emotion Recognition}

\author{Dingkang Yang$^{1,2}$ $\quad$
        Zhaoyu Chen$^{1}$ $\quad$
        Yuzheng Wang$^{1}\quad$
        Shunli Wang$^{1}\quad$
        Mingcheng Li$^{1}\quad$ 
        Siao Liu$^{1}\quad$ \\
        Xiao Zhao$^{1}\quad$
        Shuai Huang$^{1}\quad$
        Zhiyan Dong$^{1}\quad$
        Peng Zhai$^{1}\quad$
        Lihua Zhang$^{1,2,3,4}$\footnotemark[4] \\ 
        $^1$Academy for Engineering and Technology, Fudan University$\quad$
        $^2$Institute of Meta-Medical, IPASS\\
        $^3$Jilin Provincial Key Laboratory of
Intelligence Science and Engineering, Changchun, China\\
$^4$Engineering Research Center of AI and Robotics, Ministry of Education, Shanghai, China\\
{\tt\small $\{$dkyang20,lihuazhang$\}$@fudan.edu.cn}
}

\maketitle


\renewcommand{\thefootnote}{\fnsymbol{footnote}} 
\footnotetext[4]{Corresponding Author.} 


\begin{abstract}
Context-Aware Emotion Recognition (CAER) is a crucial and challenging task that aims to perceive the emotional states of the target person with contextual information. Recent approaches invariably focus on designing sophisticated architectures or mechanisms to extract seemingly meaningful representations from subjects and contexts. However, a long-overlooked issue is that a context bias in existing datasets leads to a significantly unbalanced distribution of emotional states among different context scenarios. Concretely, the harmful bias is a confounder that misleads existing models to learn spurious correlations based on conventional likelihood estimation, significantly limiting the models' performance. To tackle the issue, this paper provides a causality-based perspective to disentangle the models from the impact of such bias, and formulate the causalities among variables in the CAER task via a tailored causal graph. Then, we propose a Contextual Causal Intervention Module (CCIM) based on the backdoor adjustment to de-confound the confounder and exploit the true causal effect for model training. CCIM is plug-in and model-agnostic, which improves diverse state-of-the-art approaches by considerable margins.
Extensive experiments on three benchmark datasets demonstrate the effectiveness of our CCIM and the significance of causal insight.
\end{abstract}

\section{Introduction}
\begin{figure}[t]
  \centering
  \includegraphics[width=\linewidth]{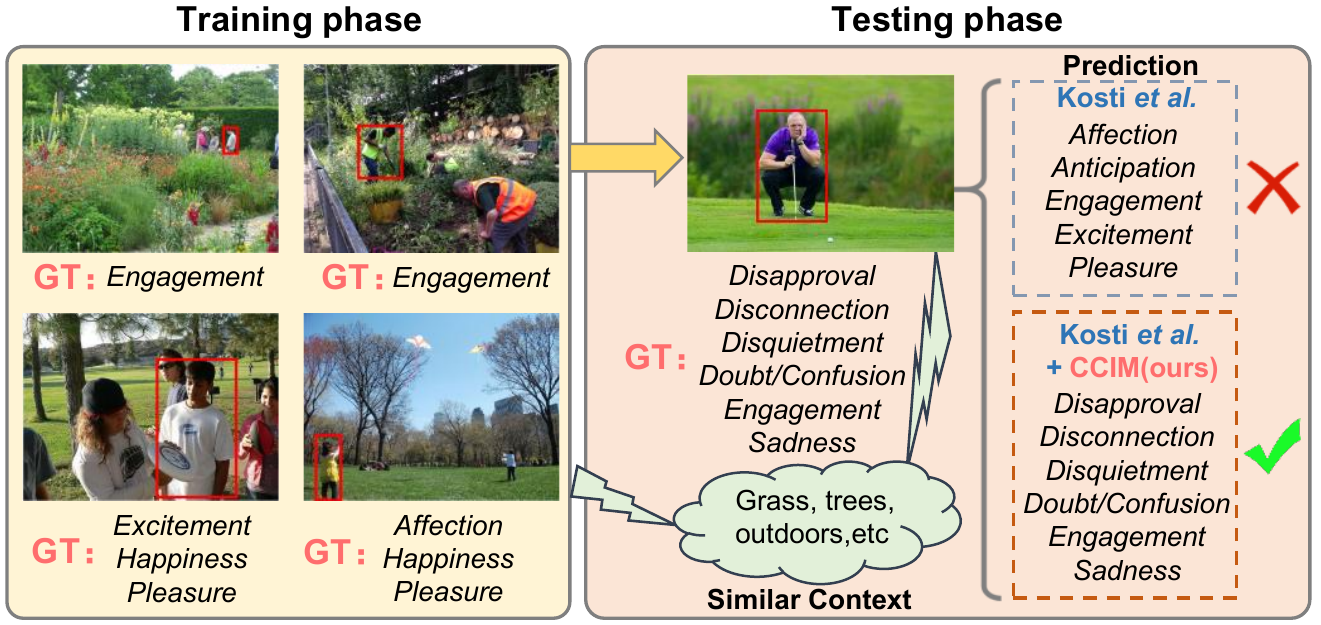}
  \caption{Illustration of the context bias in the CAER task. GT means the ground truth. Most images contain similar contexts in the training data with positive emotion categories. In this case, the model learns the spurious correlation between specific contexts and emotion categories and gives wrong results. Thanks to CCIM, the simple baseline~\cite{kosti2017emotion} achieves more accurate predictions.
  }
  \label{indro1}
\vspace{-0.4cm}
\end{figure}

As an essential technology for understanding human intentions, emotion recognition has attracted significant attention in various fields such as human-computer interaction~\cite{alnuaim2022human}, medical monitoring~\cite{meng2021hybrid}, and education~\cite{tanko2022shoelace}.
Previous works have focused on extracting multimodal emotion cues from human subjects, including facial expressions~\cite{du2021learning,farzaneh2021facial,yang2022disentangled}, acoustic behaviors~\cite{yang2022learning,ayadi2022deep,yang2022contextual}, and body postures~\cite{liu2021imigue,yang2023target},
benefiting from advances in deep learning algorithms~\cite{liu2023generalized,liu2022learning,kuang2023towards,zhu2023direct,wang2021tsa,yang2023novel,chen2022shape,chen2022towards,zhai2022robust,wang2023adversarial,wang2023explicit,wang2021survey}.
Despite the impressive improvements achieved by subject-centered approaches, their performance is limited by natural and unconstrained environments. Several examples in \Cref{indro1}~(left) show typical situations on a visual level.
Instead of well-designed visual contents, multimodal representations of subjects in wild-collected images are usually indistinguishable (\eg, ambiguous faces or gestures), which forces us to exploit complementary factors around the subject that potentially reflect emotions.

Inspired by psychological study~\cite{barrett2011context}, recent works~\cite{kosti2017emotion,lee2019context,mittal2020emoticon,zhang2019context,li2021human} have suggested that contextual information contributes to effective emotion cues for Context-Aware Emotion Recognition (CAER). The contexts are considered to include the place category, the place attributes, the objects, or the actions of others around the subject~\cite{kosti2019context}. The majority of such research
typically follows a common pipeline: (1) Obtaining the unimodal/multimodal representations of the recognized subject; (2) Building diverse contexts and extracting emotion-related representations; (3) Designing fusion strategies to combine these features for emotion label predictions.
Although existing methods have improved modestly through complex module stacking \cite{li2021human, gao2021graph,yang2022emotion} and tricks~\cite{mittal2020emoticon, hoang2021context}, they invariably suffer from a context bias of the datasets, which has long been overlooked.
Recalling the process of generating CAER datasets, different annotators were asked to label each image according to what they subjectively thought people in the images with diverse contexts were feeling~\cite{kosti2019context}.
This protocol makes the preference of annotators inevitably affect the distribution of emotion categories across contexts, thereby leading to the context bias. \Cref{indro1} illustrates how such bias confounds the predictions. Intrigued, most of the images in training data contain vegetated scenes with positive emotion categories, while negative emotions in similar contexts are almost non-existent. Therefore, the baseline~\cite{kosti2017emotion} is potentially misled into learning the spurious dependencies between context-specific features and label semantics. When given test images with similar contexts but negative emotion categories, the model inevitably infers the wrong emotional states.

 \begin{figure}[t]
  \centering
  \includegraphics[width=\linewidth]{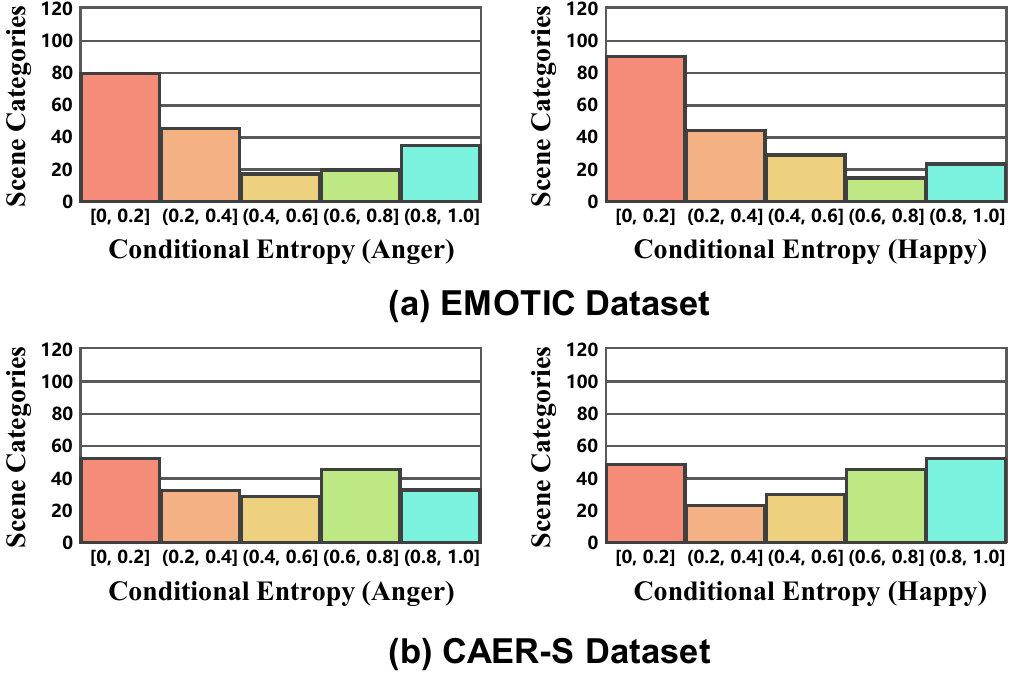}
  \caption{We show a toy experiment on the EMOTIC \cite{kosti2019context} and CAER-S \cite{lee2019context} datasets for scene categories of angry and happy emotions. More scene categories with normalized zero-conditional entropy  reveal a strong presence of the context bias.
  }
  \label{indro2}
\vspace{-0.39cm}
\end{figure}

More intrigued, a toy experiment is performed to verify the strong bias in CAER datasets. 
This test aims to observe how well emotions correlate with contexts (\eg, scene categories).
Specifically, we employ the ResNet-152~\cite{he2016deep} pre-trained on Places365~\cite{zhou2017places} to predict scene categories from images with three common emotion categories (\ie, ``anger'', ``happy'', and ``fear'') across two datasets.
The top 200 most frequent scenes from each emotion category are selected, and the normalized conditional entropy of each scene category across the positive and negative set of a specific emotion is computed~\cite{panda2018contemplating}.
While analyzing correlations between scene contexts and emotion categories in \Cref{indro2} (\eg, ``anger'' and ``happy''), we find that more scene categories with the zero conditional entropy are most likely to suggest the significant context bias in the datasets, as it shows the presence of these scenes only in the positive or negative set of emotions.
Concretely, for the EMOTIC dataset \cite{kosti2019context}, about 40\% of scene categories for anger have zero conditional entropy while about 45\% of categories for happy (\ie, happiness) have zero conditional entropy. As an intuitive example, most party-related scene contexts are present in the samples with the happy category and almost non-existent in the negative categories.
\emph{These observations confirm the severe context bias in CAER datasets, leading to distribution gaps in emotion categories across contexts and uneven visual representations}.

\begin{figure}[t]
  \centering
  \includegraphics[width=\linewidth]{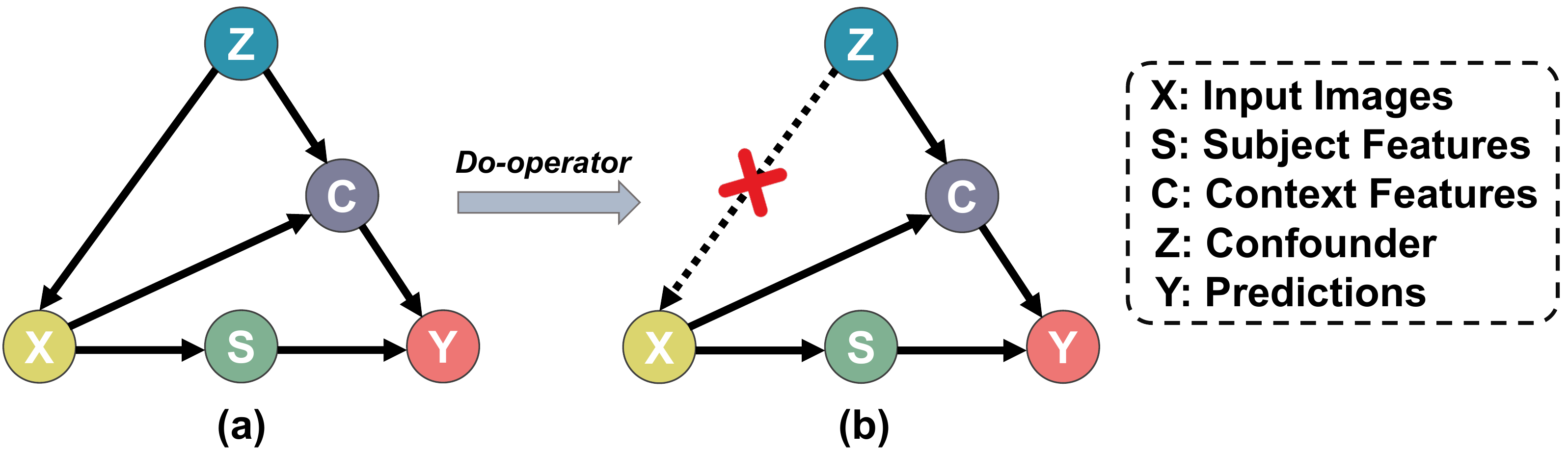}
  \caption{Illustration of our CAER causal graph. (a) The conventional likelihood $P(\bm{Y}|\bm{X})$. (b) The causal
intervention $P(\bm{Y}|do(\bm{X}))$.
  }
  \label{indro3}
\vspace{-0.3cm}
\end{figure}

Motivated by the above observation, we attempt to embrace causal inference~\cite{pearl2009causal} to reveal the culprit that poisons the CAER models, rather than focusing on beating them. As a revolutionary scientific paradigm that facilitates models toward unbiased prediction, the most important challenge in applying classical causal inference to the modern CAER task is how to reasonably depict true causal effects and identify the task-specific dataset bias. To this end, this paper attempts to address the challenge and rescue the bias-ridden models by drawing on human instincts, \ie, looking for the causality behind any association. Specifically, we present a causality-based bias mitigation strategy. We first formulate the procedure of the CAER task via a proposed causal graph.
In this case, the harmful \textbf{context bias} in datasets is essentially an unintended \textbf{confounder} that misleads the models to learn the spurious correlation between similar contexts and specific emotion semantics.
From \Cref{indro3}, we disentangle the causalities among the input images $\bm{X}$, subject features $\bm{S}$, context features $\bm{C}$, confounder $\bm{Z}$, and predictions $\bm{Y}$.
Then, we propose a simple yet effective Contextual Causal Intervention Module (CCIM) to achieve context-deconfounded training and use the $do$-calculus $P(\bm{Y}|do(\bm{X}))$ to calculate the true causal effect, which is fundamentally different from the conventional likelihood $P(\bm{Y}|\bm{X})$.
CCIM is plug-in and model-agnostic, with the backdoor adjustment~\cite{glymour2016causal} to de-confound the confounder and eliminate the impact of the context bias.
We comprehensively evaluate the effectiveness and superiority of CCIM on three standard and biased CAER datasets. Numerous experiments and analyses demonstrate that CCIM can significantly and consistently improve existing baselines, achieving a new state-of-the-art (SOTA).

The main contributions can be summarized as follows:

\begin{itemize}
\item To our best knowledge, we are the first to investigate the adverse context bias of the datasets in the CAER task from the causal inference perspective and identify that such bias is a confounder, which misleads the models to learn the spurious correlation.

\item We propose CCIM, a plug-in contextual causal intervention module, which could be inserted into most CAER models to remove the side effect caused by the confounder and facilitate a fair contribution of diverse contexts to emotion understanding.

\item Extensive experiments on three standard CAER datasets show that the proposed CCIM can facilitate existing models to achieve unbiased predictions.
\end{itemize}

\section{Related Work}
\noindent \textbf{Context-Aware Emotion Recognition.}
As a promising task, Context-Aware Emotion Recognition (CAER) not only draws on human subject-centered approaches~\cite{bhattacharya2020step,yang2022learning,yang2022disentangled} to perceive emotion via the face or body, but also considers the emotion cues provided by background contexts in a joint and boosting manner. 
Existing CAER models invariably extract multiple representations from these two sources and then perform feature fusion to make the final prediction \cite{zhang2019context,lee2019context, ruan2020context, mittal2020emoticon,li2021sequential,gao2021graph,li2021human,yang2022emotion}. For instance,
Kosti \etal \cite{kosti2017emotion} establish the EMOTIC dataset and propose a baseline Convolutional Neural Network (CNN) model that combines the body region and the whole image as the context.
Hoang \etal \cite{hoang2021context} propose an extra reasoning module to exploit the images, categories, and bounding boxes of adjacent objects in background contexts to achieve visual relationship detection.
For a deep exploration of scene context, Li \etal \cite{li2021human} present a body-object attention module to estimate the contributions of background objects and a body-part attention module to recalibrate the channel-wise body feature responses.
Although the aforementioned approaches achieve impressive improvements by exploring diverse contextual information, they all neglect the limitation on model performance caused by the context bias of the datasets. 
Instead of focusing on beating the latest SOTA, we identify the bias as a harmful confounder from a causal inference perspective and significantly improve the existing models with the proposed CCIM.

\noindent \textbf{Causal Inference.}
Causal inference is an analytical tool that aims to infer the dynamics of events under changing conditions (\eg, different treatments or external interventions)~\cite{pearl2009causal}, which has been extensively studied in economics, statistics, and psychology~\cite{varian2016causal,foster2010causal}.
Without loss of generality, causal inference follows two main ways: structured causal model~\cite{pearl2000models} and potential outcome framework~\cite{rubin2005causal},
which assist in revealing the causality rather than the superficial association among variables.
Benefiting from the great potential of the causal tool to provide unbiased estimation solutions, it has been gradually applied to various computer tasks, such as computer vision~\cite{wang2022spacenet,wang2020visual,qi2020two,tang2020unbiased,chen2022causal} and natural language processing~\cite{zhang2021biasing,qian2021counterfactual,huang2020counterfactually}.
Inspired by visual commonsense learning~\cite{wang2020visual},
to our best knowledge, this is the first investigation of the confounding effect through causal inference in the CAER task while exploiting causal intervention to interpret and address the confounding bias from contexts.

\section{Methodology}
\subsection{Causal View at CAER Task}
Firstly, we formulate a tailored causal graph to summarize the CAER framework. In particular, we follow the same graphical notation in structured causal model~\cite{pearl2000models} due to its intuitiveness and interpretability. It is a directed acyclic graph $\mathcal{G} = \{ \mathcal{N}, \mathcal{E} \}$  that can be paired with data to produce quantitative causal
estimates. The nodes $\mathcal{N}$ denote variables and the links $\mathcal{E}$ denote direct causal effects. 
As shown in Figure~\ref{indro3}, there are five variables involved in the CAER causal graph, which are input images $\bm{X}$, subject features $\bm{S}$, context features $\bm{C}$, confounder $\bm{Z}$, and predictions $\bm{Y}$.  Note that our causal graph is applicable to a variety of CAER methods, since it is highly general, imposing
no constraints on the detailed implementations. The details of the causal relationships are described below.

\noindent \textbf{$ \bm{Z} \rightarrow \bm{X} $.}
Different subjects are recorded in various contexts to produce images $\bm{X}$. 
On the one hand, the annotators make subjective and biased guesses about subjects’ emotional states and give their annotations~\cite{kosti2019context,jones1999bounded}, \eg, subjects are usually blindly assigned positive emotions in vegetation-covered contexts. On the other hand, the data nature leads to an unbalanced representation of emotions in the real world~\cite{geirhos2020shortcut}. That is, it is much easier to collect positive emotions in contexts of comfortable atmospheres than negative ones.
The context bias caused by the above situations is treated as the harmful confounder $\bm{Z}$ to establish spurious connections between similar contexts and specific emotion semantics.
For the input images $\bm{X}$, $\bm{Z}$ determines the biased content that is recorded, \ie, $ \bm{Z} \rightarrow \bm{X}$.

\noindent \textbf{$ \bm{Z} \rightarrow \bm{C}
\rightarrow \bm{Y} $.}
$\bm{C}$ represents the total context representation obtained by contextual feature extractors. $\bm{C}$ may come from the aggregation of diverse context features based on different methods. The causal path $ \bm{Z} \rightarrow \bm{C} $ 
represents the detrimental $\bm{Z}$ confounding the model to learn unreliable emotion-related context semantics of $\bm{C}$.
In this case, the impure $\bm{C}$ further affects the predictions $\bm{Y}$ of the emotion labels and can be reflected via the link $\bm{C} \rightarrow \bm{Y}$.
Although $\bm{Z}$ potentially provides priors from the training data to better estimation when the subjects' features are ambiguous, it misleads the model to capture spurious ``\emph{context-emotion}" mapping during training, resulting in biased predictions.

\noindent \textbf{$ \bm{X} \rightarrow \bm{C}
\rightarrow \bm{Y} $ \& $ \bm{X} \rightarrow \bm{S}
\rightarrow \bm{Y} $.}
$\bm{S}$ represents the total subject representation obtained by subject feature extractors.
Depending on distinct methods, $\bm{S}$ may come from the face, the body, or the integration of their features.
In CAER causal graph, we can see that the desired effect of $\bm{X}$ on $\bm{Y}$ follows from two causal paths: $ \bm{X} \rightarrow \bm{C}
\rightarrow \bm{Y} $ and $ \bm{X} \rightarrow \bm{S}
\rightarrow \bm{Y} $.
These two causal paths reflect that the CAER model estimates $\bm{Y}$ based on the context features $\bm{C}$ and subject features $\bm{S}$ extracted from the input images $\bm{X}$.
In practice, $\bm{C}$ and $\bm{S}$ are usually integrated to make the final prediction jointly, \eg,  feature concatenation~\cite{mittal2020emoticon}.

According to the causal theory~\cite{pearl2009causal}, 
the confounder $\bm{Z}$ is the common cause of the input images $\bm{X}$ and
corresponding  predictions $\bm{Y}$. 
The positive effects of context and subject features providing valuable semantics follow the causal paths $ \bm{X} \rightarrow \bm{C}/\bm{S}
\rightarrow \bm{Y} $, which we aim to achieve. Unfortunately, the confounder $\bm{Z}$  causes the negative effect of misleading the model to focus on spurious correlations instead of pure causal relationships. This adverse effect follows the backdoor causal path $\bm{X} \leftarrow \bm{Z} \rightarrow \bm{C} \rightarrow \bm{Y}$.

\subsection{Causal Intervention via Backdoor Adjustment}
In Figure~\ref{indro3}(a), existing CAER methods rely on the likelihood $P(\bm{Y}|\bm{X})$. This process is formulated by Bayes rule:
\begin{small}
\begin{equation}
P(\bm{Y}|\bm{X})=\sum_{\bm{z}}^{} P(\bm{Y}|\bm{X},\bm{S}=f_{s} (\bm{X}), \bm{C}=f_{c} (\bm{X}, \bm{z})) P(\bm{z}|\bm{X}),
\label{one}
\end{equation}
\end{small}
where $f_{s} (\cdot)$ and $f_{c} (\cdot)$ are two generalized encoding functions that obtain the total $\bm{S}$ and $\bm{C}$, respectively. The confounder $\bm{Z}$ introduces the observational bias via $ P(\bm{z}|\bm{X})$.
To address the confounding effect brought by $\bm{Z}$ and make the model rely on pure $\bm{X}$ to estimate $\bm{Y}$, an intuitive idea is to intervene $\bm{X}$ and force each context semantics to contribute to the emotion prediction fairly.
The process can be viewed as conducting a
randomized controlled experiment by collecting images of subjects with any emotion in any context. However, this intervention is impossible due to the infinite number of images that combine various subjects and contexts in the real world. To solve this, we stratify $\bm{Z}$ based on the backdoor adjustment~\cite{pearl2009causal} to achieve causal intervention $P(\bm{Y}|do(\bm{X}))$ and block the backdoor path between $\bm{X}$ and $\bm{Y}$, where $do$-calculus is an effective approximation for the imaginative intervention \cite{glymour2016causal}. Specifically,
we seek the effect of stratified contexts and then estimate the average causal
effect by computing a weighted average based on the proportion of samples containing different context prototypes in the training data. In \Cref{indro3}(b), the causal path from $\bm{Z}$ to $\bm{X}$ is cut-off, and the model will approximate causal intervention $P(\bm{Y}|do(\bm{X}))$ rather than spurious association $P(\bm{Y}|\bm{X})$.
By applying the Bayes rule on the new graph, \cref{one} with the intervention is formulated as:
\begin{small}
\begin{equation}
P(\bm{Y}|do(\bm{X}))=\sum_{\bm{z}}^{} P(\bm{Y}|\bm{X},\bm{S}=f_{s} (\bm{X}), \bm{C}=f_{c} (\bm{X}, \bm{z})) P(\bm{z}).
\label{two}
\end{equation}
\end{small}
As $\bm{z}$ is no longer affected by $\bm{X}$,
the intervention intentionally forces $\bm{X}$ to incorporate
every $\bm{z}$ fairly into the predictions of $\bm{Y}$, subject to the proportion of each $\bm{z}$ in the whole.

 \begin{figure}[t]
  \centering
  \includegraphics[width=\linewidth]{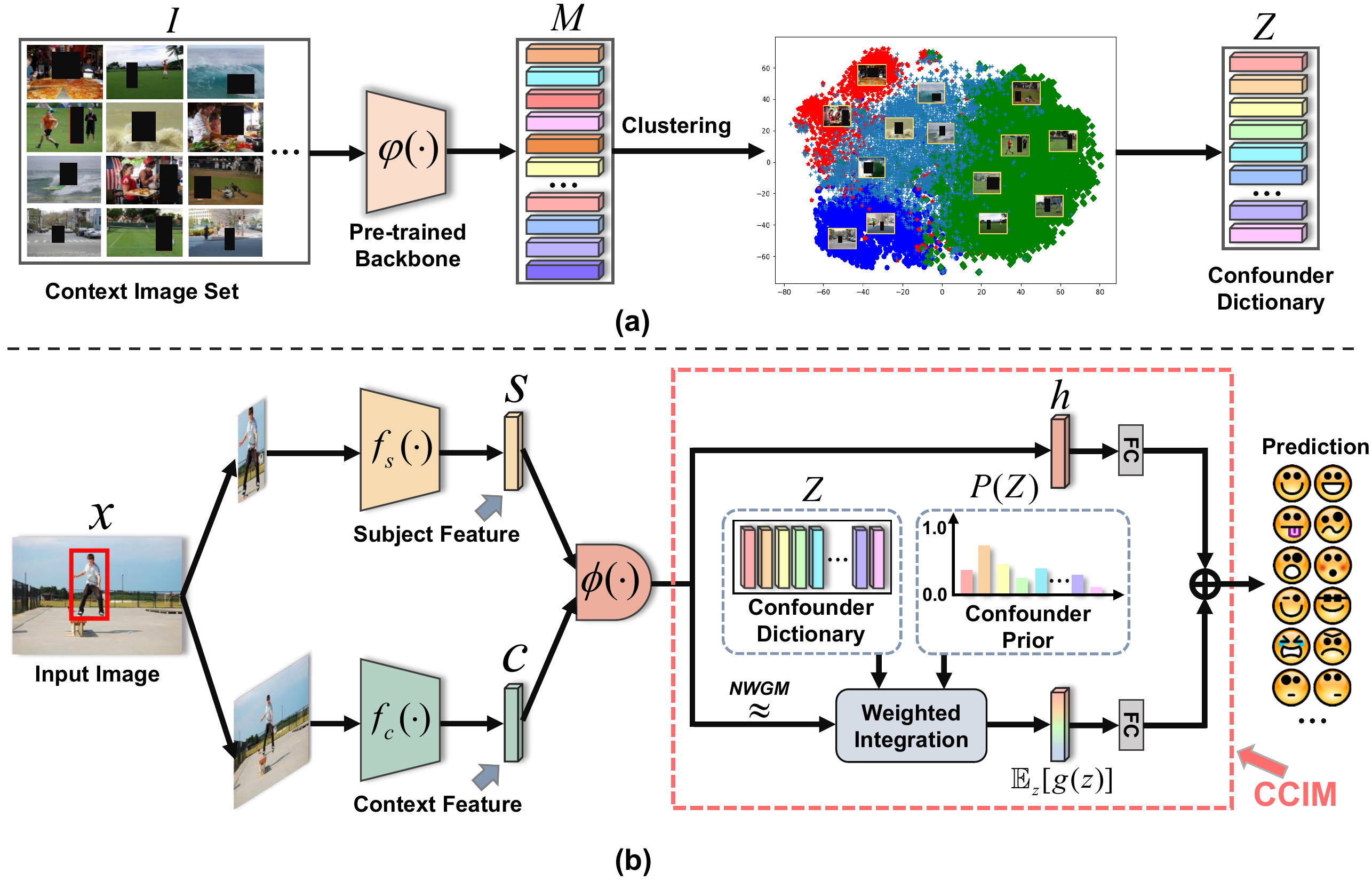}
  \caption{(a) The generation process of the confounder dictionary $\bm{Z}$. (b) A general pipeline for the context-deconfounded training. The red dotted box shows the core component that achieves the powerful approximation to causal intervention: our CCIM.
  }
  \label{art}
 \vspace{-0.5cm}
\end{figure}
\subsection{Context-Deconfounded Training with  CCIM} 
To implement the theoretical and imaginative intervention in \cref{two}, we propose a Contextual Causal Intervention Module (CCIM) to achieve the context-deconfounded training for the models.
From a general pipeline of the CAER task illustrated in \Cref{art}(b), CCIM is inserted in a plug-in manner after the original integrated feature of existing methods. Then, the output of CCIM performs predictions after passing the final task-specific classifier. The implementation of CCIM is described below.

\noindent \textbf{Confounder Dictionary.}
Since the number of contexts is large in the real world and there is no ground-truth contextual information in the training set,
we approximate it as a stratified confounder dictionary $\bm{Z}= [\bm{z}_1,\bm{z}_2, \dots,\bm{z}_{N} ]$,
where $N$ is a hyperparameter representing the size, and each $\bm{z}_i \in \mathbb{R}^{d}$ represents a context prototype.
As shown in \Cref{art}(a), we first mask the target subject in each training image based on the subject's bounding box to generate the context image set $I$.
Subsequently, the image set $I$ is fed to the pre-trained backbone network $\varphi(\cdot) $ to obtain the context feature set $\bm{M}=\{\bm{m}_{k} \in \mathbb{R}^{d}  \}^{N_{m}}_{k=1} $ , where $N_m$ is the number of training samples. 
To compute context prototypes, we use the K-Means++ with principle component analysis to learn $\bm{Z}$ so that each $\bm{z}_i$ represents a form of context cluster. Each cluster $\bm{z}_i$ is set to the average feature of each cluster in the K-Means++, \ie, $\bm{z}_i = \frac{1}{N_i}  {\sum_{j=1}^{N_i}} \bm{m}^{i} _j $, where $N_i$ is the number of context features in the $i$-th cluster.

\begin{table*}[t]
\setlength{\tabcolsep}{7pt}
\centering
\resizebox{\linewidth}{!}{%
\begin{tabular}{llcccccccccc}
\toprule
\multicolumn{2}{l}{\multirow{2}{*}{\textbf{Category}}} & \multirow{2}{*}{EMOT-Net~\cite{kosti2017emotion}} & \multirow{2}{*}{\begin{tabular}[c]{@{}c@{}}EMOT-Net\\ +~\textbf{CCIM}\end{tabular}} & \multirow{2}{*}{GCN-CNN~\cite{zhang2019context}} & \multirow{2}{*}{\begin{tabular}[c]{@{}c@{}}GCN-CNN\\ +~\textbf{CCIM}\end{tabular}} & \multirow{2}{*}{CAER-Net~\cite{lee2019context}} & \multirow{2}{*}{\begin{tabular}[c]{@{}c@{}}CAER-Net\\ +~\textbf{CCIM}\end{tabular}} & \multirow{2}{*}{RRLA~\cite{li2021human}} & \multirow{2}{*}{VRD~\cite{hoang2021context} } &  \multirow{2}{*}{EmotiCon~\cite{mittal2020emoticon}} & \multirow{2}{*}{\begin{tabular}[c]{@{}c@{}}EmotiCon\\ +~\textbf{CCIM}\end{tabular}} \\
\multicolumn{2}{l}{}                          &                             &                                                                             &                          &                                                                          &                           &                                                                           &                          &                         &                           &                                                                           \\ \midrule
\multicolumn{2}{l}{Affection}   & 26.47       & 34.87    & 47.52     & 36.18  & 22.36    & 23.08                                          & 37.93                    & 44.48                   & 38.55                     & 40.77                                                                     \\
\multicolumn{2}{l}{Anger}                     & 11.24                       & 13.05                                                                        & 11.27                    & 12.53                                                                    & 12.88                     & 12.99                                                                     & 13.73                    & 30.71                   & 14.69                     & 15.48                                                                     \\
\multicolumn{2}{l}{Annoyance}                 & 15.26                       & 18.04                                                                       & 12.33                    & 13.73                                                                    & 14.42                     & 15.28                                                                     & 20.87                    & 26.47                   & 24.68                     & 24.47                                                                     \\
\multicolumn{2}{l}{Anticipation}              & 57.31                       & 94.19                                                             & 63.2                     & 92.32                                                           & 52.85                     & 90.03                                                            & 61.08                    & 59.89                   & 60.73                     & 95.15                                                            \\
\multicolumn{2}{l}{Aversion}                  & 7.44                        & 13.41                                                              & 6.81                     & 15.41                                                          & 3.26                      & 12.96                                                            & 9.61                     & 12.43                   & 11.33                     & 19.38                                                            \\
\multicolumn{2}{l}{Confidence}                & 80.33                       & 74.9                                                                        & 74.83                    & 75.01                                                                    & 72.68                     & 73.24                                                                     & 80.08                    & 79.24                   & 68.12                     & 75.81                                                                     \\
\multicolumn{2}{l}{Disapproval}               & 16.14                       & 19.87                                                                       & 12.64                    & 14.45                                                                    & 15.37                     & 16.38                                                                     & 21.54                    & 24.54                   & 18.55                     & 23.65                                                                     \\
\multicolumn{2}{l}{Disconnection}             & 20.64                       & 27.72                                                                       & 23.17                    & 30.52                                                                    & 22.01                     & 23.39                                                                     & 28.32                    & 34.24                   & 28.73                     & 31.93                                                                     \\
\multicolumn{2}{l}{Disquietment}              & 19.57                       & 19.12                                                                       & 17.66                    & 20.85                                                                    & 10.84                     & 18.1                                                                      & 22.57                    & 24.23                   & 22.14                     & 26.84                                                                     \\
\multicolumn{2}{l}{Doubt/Confusion}           & 31.88                       & 19.35                                                                       & 19.67                    & 20.43                                                                    & 26.07                     & 17.66                                                                     & 33.5                     & 25.42                   & 38.43                     & 34.28                                                                     \\
\multicolumn{2}{l}{Embarrassment}             & 3.05                        & 6.23                                                               & 1.58                     & 9.21                                                            & 1.88                      & 5.86                                                             & 4.16                     & 4.26                    & 10.31                     & 16.73                                                           \\
\multicolumn{2}{l}{Engagement}                & 86.69                       & 88.93                                                                       & 87.31                    & 96.88                                                                    & 73.71                     & 70.04                                                                     & 88.12                    & 88.71                   & 86.23                     & 97.41                                                                     \\
\multicolumn{2}{l}{Esteem}                    & 17.86                       & 21.69                                                              & 12.05                    & 22.72                                                           & 15.38                     & 16.67                                                            & 20.5                     & 17.99                   & 25.75                     & 27.44                                                            \\
\multicolumn{2}{l}{Excitement}                & 78.05                       & 73.81                                                                       & 72.68                    & 73.21                                                                    & 70.42                     & 71.08                                                                     & 80.11                    & 74.21                   & 80.75                     & 81.59                                                                     \\
\multicolumn{2}{l}{Fatigue}                   & 8.87                        & 9.96                                                                        & 12.93                    & 12.66                                                                    & 6.29                      & 9.73                                                                      & 17.51                    & 22.62                   & 19.35                     & 15.53                                                                     \\
\multicolumn{2}{l}{Fear}                      & 15.7                        & 9.04                                                                        & 6.15                     & 10.31                                                                    & 7.47                      & 6.61                                                                      & 15.56                    & 13.92                   & 16.99                     & 15.37                                                                     \\
\multicolumn{2}{l}{Happiness}                 & 58.92                       & 78.09                                                             & 72.9                     & 75.64                                                           & 53.73                     & 62.34                                                            & 76.01                    & 83.02                   & 80.45                     & 83.55                                                            \\
\multicolumn{2}{l}{Pain}                      & 9.46                        & 14.71                                                              & 8.22                     & 15.36                                                           & 8.16                      & 9.43                                                             & 14.56                    & 16.68                   & 14.68                     & 17.76                                                            \\
\multicolumn{2}{l}{Peace}                     & 22.35                       & 22.79                                                                       & 30.68                    & 23.88                                                                    & 19.55                     & 20.21                                                                     & 26.76                    & 28.91                   & 35.72                     & 38.94                                                                     \\
\multicolumn{2}{l}{Pleasure}                  & 46.72                       & 46.59                                                                       & 48.37                    & 45.52                                                                    & 34.12                     & 35.37                                                                     & 55.64                    & 55.47                   & 67.31                     & 64.57                                                                     \\
\multicolumn{2}{l}{Sadness}                   & 18.69                       & 17.47                                                                       & 23.9                     & 22.08                                                                    & 17.75                     & 13.24                                                                     & 30.8                     & 42.87                   & 40.26                     & 45.63                                                                     \\
\multicolumn{2}{l}{Sensitivity}               & 9.05                        & 7.91                                                                        & 4.74                     & 8.02                                                                     & 6.94                      & 4.74                                                                      & 9.59                     & 15.89                   & 13.94                     & 17.04                                                                     \\
\multicolumn{2}{l}{Suffering}                 & 17.67                       & 15.35                                                                       & 23.71                    & 18.45                                                                    & 14.85                     & 11.89                                                                     & 30.7                     & 46.23                   & 48.05                     & 21.52                                                                     \\
\multicolumn{2}{l}{Surprise}                  & 22.38                       & 13.12                                                                       & 8.44                     & 13.93                                                                    & 17.46                     & 11.7                                                                      & 17.92                    & 16.27                   & 19.6                      & 26.81                                                                     \\
\multicolumn{2}{l}{Sympathy}                  & 15.23                       & 32.6                                                              & 19.45                    & 33.95                                                           & 14.89                     & 28.59                                                            & 15.26                    & 15.37                   & 16.74                     & 47.6                                                             \\
\multicolumn{2}{l}{Yearning}                  & 9.22                        & 10.08                                                                        & 9.86                     & 11.58                                                                    & 4.84                      & 8.61                                                                     & 10.11                    & 10.04                   & 15.08                     & 12.25                                                                     \\ \midrule
\multicolumn{2}{l}{mAP}              & $27.93^{\dagger}$                       & $\textbf{30.88}^{\dagger}$  (${ \uparrow~2.95 }$ )                                                            & $28.16^{\dagger}$                    & $\textbf{31.72}^{\dagger}$ (${ \uparrow~3.56 }$ )                                                            & $23.85^{\dagger}$                     & $\textbf{26.51}^{\dagger}$  (${ \uparrow~2.66 }$ )                                                           & $32.41^{\ast}$                    & $35.16^{\ast}$                    & $35.28^{\dagger}$                     & $\textbf{39.13}^{\dagger}$  (${ \uparrow~3.85 }$ )                                                                     \\ \bottomrule
\end{tabular}
}
\vspace{-5pt}
\caption{Average precision (\%) of different methods for each emotion category on the EMOTIC dataset.  $\ast$: results from the original reports. $\dagger$: results from implementation. The footnotes $\ast$ and $\dagger$ of \Cref{groupwalk,caer} follow the same interpretation.}
\label{emotic}
\vspace{-0.4cm}
\end{table*}

\noindent \textbf{Instantiation of the Proposed CCIM.}
Since the calculation of $P(\bm{Y}|do(\bm{X}))$ requires multiple forward passes of all $\bm{z}$, the computational overhead is expensive.
To reduce the computational cost, we apply the Normalized Weighted Geometric Mean (NWGM)~\cite{xu2015show} to approximate the above expectation at the feature level as:
\begin{small}
\begin{equation}
P(\bm{Y}|do(\bm{X})) \approx P(\bm{Y}|\bm{X},\bm{S}=f_{s} (\bm{X}), \\
\bm{C}= \sum_{\bm{z}}^{}f_{c} (\bm{X}, \bm{z}) P(\bm{z})).
\label{three}
\end{equation}
\end{small}
Inspired by~\cite{wang2020visual}, we parameterize a network model to approximate the above conditional probability of \cref{three} as follows:
\begin{small}
\begin{equation}
P(\bm{Y}|do(\bm{X})) = \bm{W}_{h}\bm{h}+\bm{W}_{g}\mathbb{E}_{\bm{z}}[g(\bm{z})],
\label{four}
\end{equation}
\end{small}
where $\bm{W}_{h}\in \mathbb{R}^{d_{m} \times d_{h}}$ and $\bm{W}_{g}\in \mathbb{R}^{d_{m} \times d}$ are the learnable parameters, and $\bm{h}=\phi (\bm{s}, \bm{c})  \in \mathbb{R}^{d_{h} \times 1}$. $\phi (\cdot)$ is a fusion strategy (\emph{e.g}., concatenation) that integrates $\bm{s}$ and $\bm{c}$ into the joint representation $\bm{h}$.
Note that the above approximation is reasonable, because
the effect on $\bm{Y}$ comes from $\bm{S}$, $\bm{C}$, and the confounder $\bm{Z}$.
Immediately, we approximate $\mathbb{E}_{\bm{z}}[g(\bm{z})]$ as a weighted integration of all context prototypes:
\begin{small}
\begin{equation}
\mathbb{E}_{\bm{z}}[g(\bm{z})]=
\sum_{i=1}^{N} \lambda_{i} \bm{z}_{i} P(\bm{z}_{i}),
\label{five}
\end{equation}
\end{small}
where $\lambda_{i}$ is a weight coefficient that measures the importance of each $\bm{z}_{i}$ after interacting with the origin feature $\bm{h}$, and $P(\bm{z}_{i})= \frac{N_i}{N_m}$.
In practice, we provide two implementations of $\lambda_{i}$: dot product attention and additive attention:
\begin{small}
\begin{align}
   \text{Dot Product}:  \lambda_{i} &= softmax(\frac{(\bm{W}_{q} \bm{h})^{T} (\bm{W}_{k} \bm{z}_{i})}
    {\sqrt{d}} ),
    \label{six}
    \\
   \text{Additive}:  \lambda_{i} &= softmax(\bm{W}_{t}^{T}
    \cdot
    Tanh(\bm{W}_{q}\bm{h}+\bm{W}_{k} \bm{z}_{i})),
    \label{seven}
\end{align}
\end{small}
where $\bm{W}_{t} \in \mathbb{R}^{d_{n} \times 1}$, $\bm{W}_{q}  \in \mathbb{R}^{d_{n} \times d_{h}}$, and $\bm{W}_{k}  \in \mathbb{R}^{d_{n} \times d}$ are mapping matrices.

\begin{table*}[t]
\centering
\resizebox{\linewidth}{!}{%
\begin{tabular}{lcccccccc}
\toprule
\multirow{2}{*}{\textbf{Category}} & \multirow{2}{*}{EMOT-Net~\cite{kosti2017emotion}} & \multirow{2}{*}{\begin{tabular}[c]{@{}c@{}}EMOT-Net\\  + \textbf{CCIM}\end{tabular}} & \multirow{2}{*}{GNN-CNN~\cite{zhang2019context}} & \multirow{2}{*}{\begin{tabular}[c]{@{}c@{}}GNN-CNN\\ + \textbf{CCIM}\end{tabular}} & \multirow{2}{*}{CAER-Net~\cite{lee2019context}} & \multirow{2}{*}{\begin{tabular}[c]{@{}c@{}}CAER-Net\\ + \textbf{CCIM}\end{tabular}} & \multirow{2}{*}{EmotiCon~\cite{mittal2020emoticon}} & \multirow{2}{*}{\begin{tabular}[c]{@{}c@{}}EmotiCon \\ + \textbf{CCIM}\end{tabular}} \\
                          &                           &                                                                             &                          &                                                                           &                           &                                                                            &                           &                                                                             \\ \midrule
Angry                     & 57.65                     & 62.41                                                                       & 51.92                    & 54.07                                                                     & 45.18                     & 50.43                                                                      & 68.85                     & 75.93                                                                       \\
Happy                     & 71.32                     & 75.68                                                                       & 63.37                    & 70.25                                                                     & 56.59                     & 60.71                                                                      & 72.31                     & 79.15                                                                       \\
Neutral                   & 43.1                      & 41.03                                                                       & 40.26                    & 39.49                                                                     & 39.32                     & 37.84                                                                      & 50.34                     & 48.66                                                                       \\
Sad                       & 61.24                     & 63.84                                                                       & 58.15                    & 61.85                                                                     & 52.96                     & 54.06                                                                      & 70.8                      & 73.48                                                                       \\ \midrule
mAP                       & $58.33^{\dagger}$                   & $\textbf{60.74}^{\dagger}$ (${ \uparrow~2.41 }$ )                                                              & $53.43^{\dagger}$                    & $\textbf{56.42}^{\dagger}$ (${ \uparrow~2.99 }$ )                                                          & $48.51^{\dagger}$                      & $\textbf{50.76 }^{\dagger}$ (${ \uparrow~2.25 }$ )                                                             & $65.58^{\dagger}$                      & $\textbf{69.31 }^{\dagger}$ (${ \uparrow~3.73 }$ )                                                             \\ \bottomrule
\end{tabular}
}
\vspace{-5pt}
\caption{Average precision (\%) of different methods for each emotion category on the GroupWalk dataset.}
\label{groupwalk}
\vspace{-0.5cm}
\end{table*}

\begin{table}[t]
\vspace{-10pt}
\centering

\resizebox{\linewidth}{!}{%
\begin{tabular}{clcclc}
\multicolumn{1}{l}{}        &       & \multicolumn{1}{l}{}                & \multicolumn{1}{l}{}        &       & \multicolumn{1}{l}{} \\
\multicolumn{1}{l}{}        &       & \multicolumn{1}{l}{}                & \multicolumn{1}{l}{}        &       & \multicolumn{1}{l}{} \\ \toprule
\multicolumn{2}{c}{\textbf{Methods}}         & \multicolumn{1}{c|}{\textbf{Accuracy}~(\%)}       & \multicolumn{2}{c}{\textbf{Methods}}         & \textbf{Accuracy}~(\%)             \\ \midrule
\multicolumn{2}{c}{CAER-Net~\cite{lee2019context}}        & \multicolumn{1}{c|}{$73.47^{\dagger}$}          & \multicolumn{2}{c}{EmotiCon~\cite{mittal2020emoticon}}        & $88.65^{\dagger}$                \\
\multicolumn{2}{c}{CAER-Net + \textbf{CCIM}} & \multicolumn{1}{c|}{$\textbf{74.81}^{\dagger}$ (${ \uparrow~1.34 }$ ) } & \multicolumn{2}{c}{EmotiCon + \textbf{CCIM}} & $\textbf{91.17}^{\dagger}$ (${ \uparrow~2.52 }$ )       \\
\multicolumn{2}{c}{EMOT-Net~\cite{kosti2017emotion}}        & \multicolumn{1}{c|}{$74.51^{\dagger}$}          & \multicolumn{2}{c}{SIB-Net \cite{li2021sequential}}         & $74.56^{\ast}$                \\
\multicolumn{2}{c}{EMOT-Net + \textbf{CCIM}} & \multicolumn{1}{c|}{$\textbf{75.82}^{\dagger}$ (${ \uparrow~1.31 }$ )} & \multicolumn{2}{c}{GRERN \cite{gao2021graph}}           & $81.31^{\ast}$               \\
\multicolumn{2}{c}{GNN-CNN~\cite{zhang2019context}}         & \multicolumn{1}{c|}{$77.21^{\dagger}$}          & \multicolumn{2}{c}{RRLA~\cite{li2021human}}            & $84.82^{\ast}$                \\
\multicolumn{2}{c}{GNN-CNN + \textbf{CCIM}}  & \multicolumn{1}{c|}{$\textbf{78.66}^{\dagger}$ (${ \uparrow~1.45 }$ ) } & \multicolumn{2}{c}{VRD~\cite{hoang2021context}}             & $90.49^{\ast}$                \\ \bottomrule
\end{tabular}
}
\vspace{-5pt}
\caption{Emotion classification accuracy (\%) of different methods on
the CAER-S dataset.}
\label{caer}
\vspace{-0.5cm}
\end{table}

\section{Experiments}
\subsection{Datasets and Evaluation  Metrics}
\noindent \textbf{Datasets.} Our experiments are conducted on three standard datasets for the CAER task, namely EMOTIC~\cite{kosti2019context}, CAER-S~\cite{lee2019context}, and GroupWalk~\cite{mittal2020emoticon} datasets.
\textbf{EMOTIC} contains 23,571 images of 34,320 annotated subjects in uncontrolled environments.
The annotation of these images contains the bounding boxes of the target subjects' body regions and 26 discrete emotion categories.
The standard partitioning of the dataset is 70\% training set, 10\% validation set, and 20\% testing test.
\textbf{CAER-S} includes 70k static images extracted
from video clips of 79 TV shows to predict emotional states. These images are randomly split into training (70\%), validation (10\%), and testing (20\%) images.
These images are annotated with 7 emotion categories: Anger, Disgust, Fear, Happy, Sad, Surprise, and Neutral. 
\textbf{GroupWalk} consists of 45 videos that were captured using stationary cameras in 8 real-world settings. The annotations consist of the following discrete labels: Angry, Happy, Neutral, and Sad. The dataset is split into 85\% training set and 15\% testing set.

\noindent \textbf{Evaluation  Metrics.}
Following \cite{kosti2017emotion,mittal2020emoticon}, we utilize the mean Average Precision (mAP) to evaluate the results on the EMOTIC and GroupWalk. For the CAER-S, the standard classification accuracy is used for evaluation.

\subsection{Model Zoo}
Limited by the fact that most methods are not open source, we select four representative models to evaluate the effectiveness of CCIM, which have different network structures and contextual exploration mechanisms. \textbf{EMOT-Net}~\cite{kosti2017emotion} is a baseline Convolutional Neural Network (CNN) model with two branches. Its distinct branches capture foreground body features and background contextual information, respectively. \textbf{GCN-CNN}~\cite{zhang2019context} utilizes different context elements to construct an affective graph and infer the affective relationship according to the Graph Convolutional Network (GCN).
\textbf{CAER-Net}~\cite{lee2019context} is a two-stream CNN model following an adaptive fusion module to reason emotions. The method focuses on the context of the entire image after hiding the face and the emotion cues provided by the facial region.
\textbf{EmotiCon}~\cite{mittal2020emoticon} introduces three context-aware streams. Besides the subject-centered multimodal extraction branch, they propose to use visual attention and depth maps to learn the scene and socio-dynamic contexts separately.
For EMOT-Net, we re-implement the model following the available code.
Meanwhile, we reproduce the results on the three datasets based on the details reported in the SOTA methods above (\ie, GCN-CNN, CAER-Net, and EmotiCon).

\subsection{Implementation Details}
\noindent \textbf{Confounder Setup.}
Firstly, except for the annotated EMOTIC, we utilize the pre-trained Faster R-CNN~\cite{ren2015faster} to detect the bounding box of the target subject for each training sample on both CAER-S and GroupWalk.
After that, the context images are generated by masking the target subjects on the training samples based on the bounding boxes.
Then, we use the ResNet-152~\cite{he2016deep} pre-trained on Places365~\cite{zhou2017places} dataset to extract the context feature set $\bm{M}$.
Each context feature $\bm{m}$ is extracted from the last pooling layer, and the hidden dimension $d$ is 2048.
The rich scene context semantics in Places365 facilitate obtaining better context prototypes from the pre-trained backbone.
In the EMOTIC, CAER-S, and GroupWalk, the default size $N$ (\ie, the number of clusters) of $\bm{Z}$ is 256, 128, and 256, respectively.

\noindent \textbf{Training Details.}
The CCIM and reproducible methods are implemented through PyTorch platform~\cite{pytorch2018pytorch}.
All models are trained on four Nvidia Tesla V100 GPUs. For a fair comparison, the training settings (\eg, loss function, batch size, learning rate strategy, etc) of these models are consistent with the details reported in their original papers.
For the implementation of our CCIM,
the hidden dimensions $d_{m}$ and $d_{n}$ are set to 128 and 256, respectively.
The output dimension $d_{h}$ of the joint feature $\bm{h}$ in the different methods is 256 (EMOT-Net), 1024 (GCN-CNN), 128 (CAER-Net), and 78 (EmotiCon).

\subsection{Comparison with State-of-the-art Methods}
We comprehensively compare the CCIM-based models with recent SOTA methods, including RRLA~\cite{li2021human}, VRD~\cite{hoang2021context}, SIB-Net \cite{li2021sequential}, and GRERN \cite{gao2021graph}. The default setting uses the dot product attention of \cref{six}.

\noindent \textbf{Results on the EMOTIC Dataset.}
In \Cref{emotic}, we observe that CCIM significantly improves existing models and achieves the  new SOTA. Specifically, the CCIM-based EMOT-Net, GCN-CNN, CAER-Net and EmotiCon improve the mAP scores by 2.95\%, 3.56\%, 2.66\%, and 3.85\%, respectively, outperforming the vanilla methods by large margins.
In this case, these CCIM-based methods achieve competitive or better performance than the recent models RRLA and VRD. We also find that CCIM greatly improves the AP scores for some categories heavily persecuted by the confounder. For instance, CCIM helps raise the results of ``Anticipation'' and ``Sympathy'' in these CAER methods by 29\%$\sim$37\% and 14\%$\sim$29\%, respectively. Due to the adverse bias effect, the performance of most models is usually poor on infrequent categories, such as ``Aversion'' (AP scores of about 3\%$\sim$12\%) and ``Embarrassment'' (AP scores of about 1\%$\sim$10\%). Thanks to CCIM, the AP scores in these two categories are achieved at about 12\%$\sim$19\% and 5\%$\sim$16\%.

\noindent \textbf{Results on the GroupWalk Dataset.}
As shown in \Cref{groupwalk}, our CCIM effectively improves the performance of EMOT-Net, GCN-CNN, CAER-Net, and EmotiCon on the GroupWalk dataset. The mAP scores for these models are increased by 2.41\%, 2.99\%, 2.25\%, and 3.73\%, respectively.

\noindent \textbf{Results on the CAER-S Dataset.}
The accuracy of different methods on the CAER-S dataset is reported in \Cref{caer}.
The performance of EMOT-Net, GCN-CNN, and CAER-Net is consistently increased by CCIM, making each context prototype contribute fairly to the emotion classification results.
These models are improved by 1.31\%, 1.45\%, and 1.34\%, respectively.
Moreover, the CCIM-based EmotiCon achieves a significant gain of 2.52\% and outperforms all SOTA methods with an accuracy of 91.17\%.

 \begin{figure}[t]
  \centering
  \includegraphics[width=0.9\linewidth]{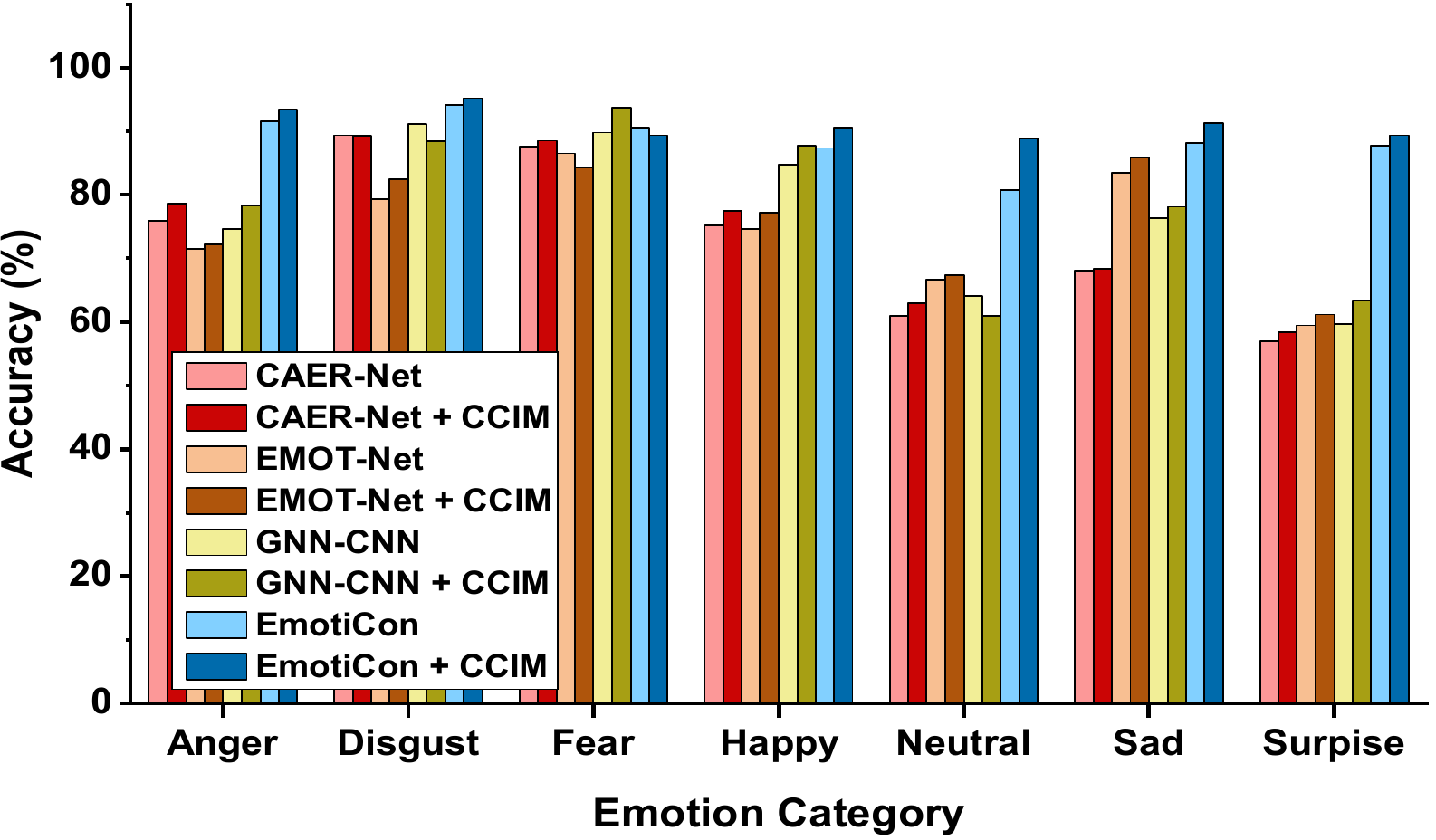}
  \caption{Emotion classification accuracy (\%) for each category of different methods on the CAER-S dataset.
  }
  \label{accuracy}
\vspace{-0.5cm}
\end{figure}

\noindent \textbf{Discussion from the Causal Perspective.}
\textbf{(\rmnum{1})} Compared to the CAER-S (average gain of 1.66\% across models), the performance improvements on the EMOTIC (average gain of 3.26\%) and GroupWalk (average gain of 2.85\%) are more significant. The potential reason is that the samples in these two datasets come from uncontrolled real-world scenarios that contain various context prototypes, such as rich scene information and agent interaction.
In this case, CCIM can more effectively eliminate spurious correlations caused by the adequately extracted confounder and provide sufficient gains. \textbf{(\rmnum{2})} Furthermore, CCIM can provide better gains for fine-grained methods of modeling context semantics. For instance, EmotiCon (average gain of 3.37\% across datasets) with two contextual feature streams significantly outperforms EMOT-Net (average gain of 2.22\%) with only one stream.
We argue that the essence of fine-grained modeling is the potential context stratification within the sample from the perspective of backdoor adjustment.
Fortunately, CCIM can better refine this stratification effect and make the models focus on contextual causal intervention across samples to measure the true causal effect. 
\textbf{(\rmnum{3})} According to \Cref{emotic,groupwalk}, and \Cref{accuracy}, while the causal intervention brings gains for most emotions across datasets, the performance of some categories shows slight improvements or even deteriorations. A reasonable explanation is that the few samples and insignificant confounding effects of these categories result in over-intervention. However, the minor sacrifice is tolerable compared to the overall superiority of our CCIM.

\subsection{Ablation Studies}
We conduct thorough ablation studies in \Cref{ablition} to evaluate the implementation of the causal intervention. 
To explore the effectiveness of CCIM when combining methods that model context semantics at different granularities, we choose the baseline EMOT-Net and SOTA EmotiCon.

\noindent \textbf{Rationality of Confounder Dictionary $\bm{Z}$.}
We first provide a random dictionary with the same size to replace the tailored confounder dictionary $\bm{Z}$, which is initialized by randomization rather than the average context features. Experimental results (3, 4) show that the random dictionary would significantly hurt the performance, proving the validity of our context prototypes. Moreover, we use the ResNet-152 pre-trained on ImageNet \cite{deng2009imagenet} to replace the default settings (1, 2) for extracting context features. The decreased results (5, 6) suggest that context prototypes based on scene semantics are more conducive to approximating the confounder than those based on object semantics. It is reasonable as scenes usually include objects, \eg, in \Cref{indro1}, ``grass'' is the child of the confounder ``vegetated scenes''.

\begin{table}[t]
\centering

\resizebox{0.95\linewidth}{!}{%
\begin{tabular}{c|cl|ccc}
\toprule
\multirow{2}{*}{ID} & \multicolumn{2}{c|}{\multirow{2}{*}{Setting}}      & EMOTIC         & CAER-S         & GroupWalk      \\ \cline{4-6} 
                    & \multicolumn{2}{c|}{}                             & mAP (\%)            & Accuracy (\%)      & mAP (\%)             \\ \midrule
\textbf{(1)}                 & \multicolumn{2}{c|}{EMOT-Net + \textbf{CCIM}}               & \textbf{30.88} & \textbf{75.82} & \textbf{60.74} \\
\textbf{(2)}                  & \multicolumn{2}{c|}{EmotiCon + \textbf{CCIM}}               & \textbf{39.13} & \textbf{91.17} & \textbf{69.31} \\ \midrule
(3)                 & \multicolumn{2}{c|}{(1) w/ Random $\bm{Z}$}              & 26.56          & 73.36          & 57.45          \\
(4)                 & \multicolumn{2}{c|}{(2) w/ Random $\bm{Z}$}              & 35.12         & 87.34          & 65.62          \\
(5)                 & \multicolumn{2}{c|}{(1) w/ ImageNet Pre-training} & 28.72          & 74.75          & 58.96          \\
(6)                 & \multicolumn{2}{c|}{(2) w/ ImageNet Pre-training} & 37.48          & 90.46          & 68.28          \\ \midrule
(7)                 & \multicolumn{2}{c|}{(1) w/ ResNet-50}             & 29.53          & 75.34          & 59.92        \\
(8)                 & \multicolumn{2}{c|}{(2) w/ ResNet-50}             & 38.86          & 90.41          & 68.85          \\
(9)                 & \multicolumn{2}{c|}{(1) w/ VGG-16}                & 28.78          & 74.95          & 59.47         \\
(10)                & \multicolumn{2}{c|}{(2) w/ VGG-16}                & 37.93         & 89.82          & 68.11          \\ \midrule
(11)                & \multicolumn{2}{c|}{(1) w/ Additive Attention}    & 30.79          & 75.64          & \textbf{60.85}        \\
(12)                & \multicolumn{2}{c|}{(2) w/ Additive Attention}    & \textbf{39.16}         & 91.08          & 69.26          \\
(13)                & \multicolumn{2}{c|}{(1) w/o $\lambda_{i}$}                    & 30.05          & 75.21          & 59.83          \\
(14)                & \multicolumn{2}{c|}{(2) w/o $\lambda_{i}$}                    & 38.53         & 89.67          & 68.75          \\
(15)                & \multicolumn{2}{c|}{(1) w/o $P(\bm{z}_{i})$}                 & 30.63          & 75.59          & 59.94          \\
(16)                & \multicolumn{2}{c|}{(2) w/o $P(\bm{z}_{i})$}                 & 39.05         & 90.06          & 69.15          \\ \midrule
(17)                & \multicolumn{2}{c|}{(1) w/o Masking Strategy}                 & 29.86          & 74.84          & 59.22          \\
(18)                & \multicolumn{2}{c|}{(2) w/o Masking Strategy}                 & 38.06         & 90.57          & 67.79          \\ \bottomrule
\end{tabular}
}
\caption{Ablation study results on all three datasets. w/ and w/o are short for with and without, respectively.}
\label{ablition}
\end{table}

 \begin{figure}[t]
  \centering
  \includegraphics[width=\linewidth]{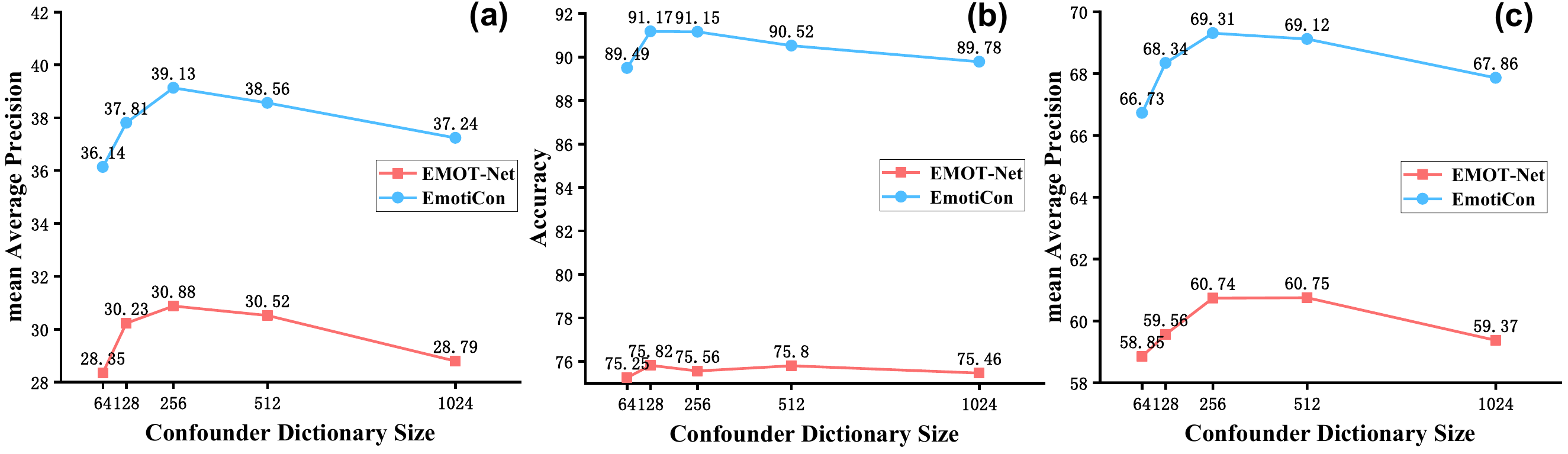}
  \caption{Ablation study results for the size $N$ of the confounder dictionary $\bm{Z}$ on three datasets. (a), (b), and (c) from the EMOTIC, CAER-S, and GroupWalk datasets, respectively.
  }
  \label{size}
\vspace{-0.5cm}
\end{figure}

\noindent \textbf{Robustness of Pre-trained Backbones.}
The experiments (7, 8, 9, 10) in Table~\ref{ablition} show that the gain from CCIM increases as more advanced pre-trained backbone networks are used, which indicates that our CCIM is not dependent on a well-chosen pre-trained backbone $\varphi(\cdot)$.

\noindent \textbf{Effectiveness of Components of $\mathbb{E}_{\bm{z}}[g(\bm{z})]$.}
First, we report the results in experiments (11, 12) using the additive attention weight $\lambda_{i}$ in 
Eq.~(\ref{seven}). The competitive performance demonstrates that both attention paradigms are meaningful and usable.
Furthermore, we evaluate the effectiveness of the weighted integration by separately removing the weights $\lambda_{i}$ and prior probabilities $P(\bm{z}_{i})$ in the $\mathbb{E}_{\bm{z}}[g(\bm{z})]$.
The decreased results (13, 14, 15, 16) suggest that depicting the importance and proportion of each confounder is indispensable for achieving effective causal intervention.

\noindent \textbf{Effect of Confounder Size.}
To justify the size $N$ of the confounder $\bm{Z}$, we set $N$ to 64, 128, 256, 512, and 1024 on all datasets separately to perform experiments.
The results in \Cref{size} show that selecting the suitable size $N$ for a dataset containing varying degrees of the harmful bias can well help the models perform de-confounded training.

 \begin{figure}[t]
  \centering
  \includegraphics[width=0.9\linewidth]{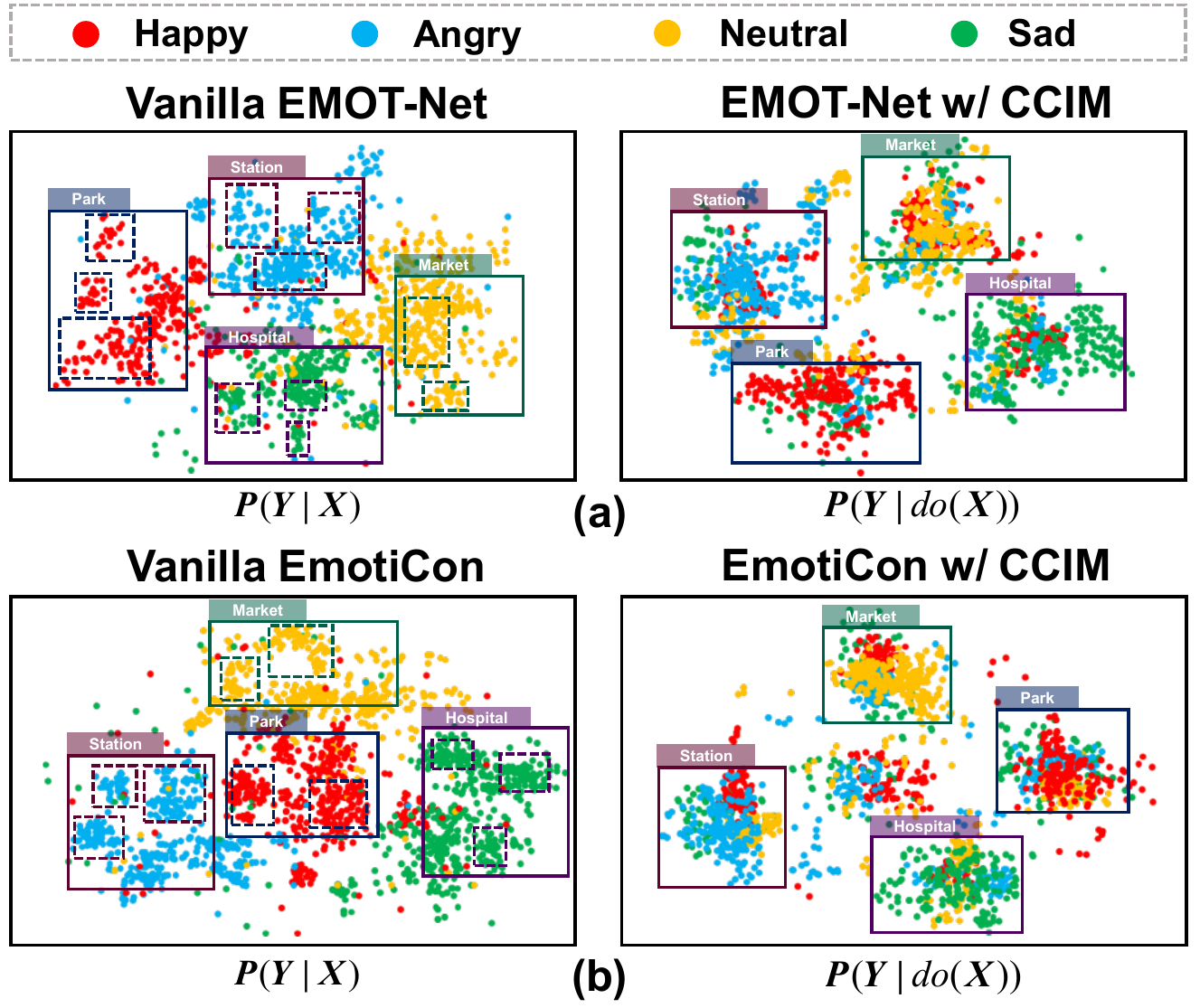}
  \caption{Visualization results of vanilla and CCIM-based EMOT-Net and EmotiCon models on the GroupWalk dataset.
  }
  \label{vis1}
  \vspace{-0.4cm}
\end{figure}

\noindent \textbf{Necessity of Masking Strategy.}
The masking strategy aims to mask the \emph{recognized subject} to learn prototype representations using pure background contexts. Note that other subjects are considered as background to provide the socio-dynamic context. The gain degradation from the experiments (17, 18) is observed when the target subject regions are not masked. It is unavoidable because the target subject-based feature attributes would impair the context-based confounder dictionary $\bm{Z}$ along the undesirable link $ \bm{S} \rightarrow \bm{Z} $, affecting causal intervention performance.
\subsection{Qualitative Results}
\noindent \textbf{Difference Between $P(\bm{Y}|\bm{X})$ and  $P(\bm{Y}|do(\bm{X}))$.}
To visually show the difference between the models approximate $P(\bm{Y}|\bm{X})$ and $P(\bm{Y}|do(\bm{X}))$, we visualize the distribution of context features learned by EMOT-Net and EmotiCon in testing samples on the GroupWalk. These sample images contain four real-world contexts, \ie, park, market, hospital, and station. \Cref{vis1} shows the following observations. In vanilla models, features with the same emotion categories usually cluster within similar context clusters (\eg, context features of the hospital with the sad category are closer), implying that biased models rely on context-specific semantics to infer emotions lopsidedly. Conversely, in the CCIM-based models, context-specific features form clusters containing diverse emotion categories. The phenomenon suggests that the causal intervention promotes models to fairly incorporate each context prototype semantics when predicting emotions, alleviating the effect of  harmful context bias.

\begin{figure}[t]
  \centering
  \includegraphics[width=0.9\linewidth]{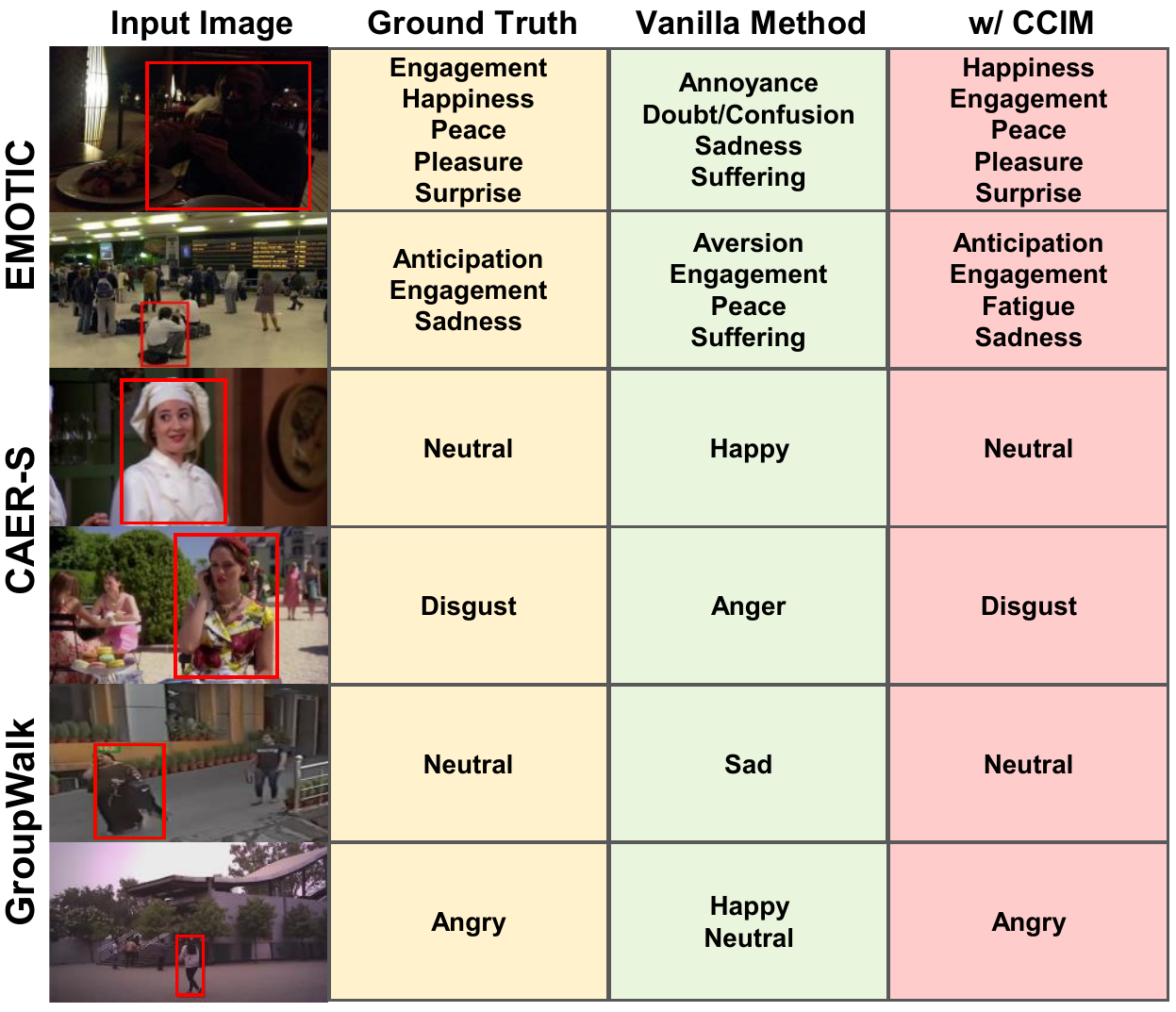}
  \caption{Qualitative results of the vanilla and CCIM-based EMOT-Net on three datasets.
  }
  \label{vis2}
\vspace{-0.4cm}
\end{figure}

\noindent \textbf{Case Study of Causal Intervention.}
In \Cref{vis2}, we select two representative examples from each dataset to show the performance of the model before and after the intervention. For instance, in the first row, the vanilla baseline is misled to predict entirely wrong results because the subjects in the dim scenes are mostly annotated with negative emotions. Thanks to causal intervention, CCIM corrects the bias in the model's prediction. Furthermore, in the fifth row, CCIM disentangles the spurious correlation between the context (``hospital entrance'') and the emotion semantics (``sad''), improving the model's performance.

\section{Conclusion}
This paper proposes a causal debiasing strategy to reduce the harmful bias of uneven distribution of emotional states across diverse contexts in the CAER task. Concretely, we disentangle the causalities among variables via a tailored causal graph and present a Contextual Causal Intervention Module (CCIM) to remove the adverse effect caused by the context bias as a confounder. Numerous experiments prove that CCIM can consistently improve existing models and promote them to a new SOTA. The model-agnostic and plug-in CCIM undoubtedly has excellent superiority over complex module stacking in previous approaches. 

\section*{Acknowledgements}
This work is supported in part by the National Key R\&D Program of China (2021ZD0113502, 2021ZD0113503), in part by the Shanghai Municipal Science and Technology Major Project (2021SHZDZX0103), and in part by the China Postdoctoral Science Foundation under Grant (BX20220071, 2022M720769).

{\small
\bibliographystyle{ieee_fullname}
\bibliography{116}
}

\end{document}